\documentclass[runningheads]{llncs}

\usepackage{graphicx}
\usepackage[ruled, linesnumbered]{algorithm2e}
\usepackage{amssymb,mathtools}
\usepackage{xcolor}
\usepackage{comment}
\usepackage{mathtools}
\usepackage[thinlines]{easytable}
\usepackage{babel,blindtext}

\usepackage{diagbox}
\usepackage{multirow}
\usepackage{longtable}
\usepackage{dblfloatfix}
\usepackage{scrextend}
\usepackage{float}
\usepackage{subcaption}
\usepackage{url}
\usepackage{amsfonts}

\usepackage{array}
\newcolumntype{L}[1]{>{\raggedright\let\newline\\\arraybackslash\hspace{0pt}}m{#1}}
\newcolumntype{C}[1]{>{\centering\let\newline\\\arraybackslash\hspace{0pt}}m{#1}}
\newcolumntype{R}[1]{>{\raggedleft\let\newline\\\arraybackslash\hspace{0pt}}m{#1}}
\usepackage{tikz,xcolor,hyperref}
\usepackage{hyperref}
\usepackage{academicons}
\definecolor{lime}{HTML}{A6CE39}
\DeclareRobustCommand{\orcidicon}{%
	\begin{tikzpicture}
	\draw[lime, fill=lime] (0,0) 
	circle [radius=0.16] 
	node[white] {{\fontfamily{qag}\selectfont \tiny ID}};
	\draw[white, fill=white] (-0.0625,0.095) 
	circle [radius=0.007];
	\end{tikzpicture}
	\hspace{-2mm}
}

\foreach \x in {A, ..., Z}{%
	\expandafter\xdef\csname orcid\x\endcsname{\noexpand\href{https://orcid.org/\csname orcidauthor\x\endcsname}{\noexpand\orcidicon}}
}

\begin{document}

\title{Combating Hostility: Covid-19 Fake News and Hostile Post Detection in Social Media}
%

\author{Omar Sharif\inst{*}\orcidA{} \and Eftekhar Hossain\inst{\textdagger}\orcidB{} \and Mohammed Moshiul Hoque\inst{*}\orcidC{}}
\institute{
\textsuperscript{*}Department of Computer Science and Engineering\\
\textsuperscript{\textdagger}Department of Electronics and Telecommunication Engineering\\
Chittagong University of Engineering and Technology, Bangladesh\\
\email{\{omar.sharif, eftekhar.hossain, moshiul\_240\}@cuet.ac.bd}}
\authorrunning{Sharif et al.}
\titlerunning{Covid-19 Fake News and Hostile Post Detection in Social Media}
%

%
\maketitle              
\begin{abstract}
This paper illustrates a detail description of the system and its results that developed as a part of the participation at CONSTRAINT shared task in AAAI-2021. The shared task comprises two tasks: a) COVID19 fake news detection in English b) Hostile post detection in Hindi. Task-A is a binary classification problem with fake and real class, while task-B is a multi-label multi-class classification task with five hostile classes (i.e. defame, fake, hate, offence, non-hostile). Various techniques are used to perform the classification task, including SVM, CNN, BiLSTM, and CNN+BiLSTM with tf-idf and Word2Vec embedding techniques. Results indicate that SVM with tf-idf features achieved the highest 94.39\% weighted $f_1$ score on the test set in task-A. Label powerset SVM with n-gram features obtained the maximum coarse-grained and fine-grained $f_1$ score of 86.03\% and 50.98\% on the task-B test set respectively.

\end{abstract}
\section{Introduction}
In recent years there has been a phenomenal surge in the number of users in social media platforms (i.e. Facebook, Twitter) to communicate, publish content, and express their opinions. The swelling number of users has resulted in the generation of the countless amount of posts on social media platforms \cite{zeitel2014social}. Although communication proliferated via the social media platforms, they also create space for anti-social and unlawful activities such as disseminating preposterous information, rumour, bullying, harassing, stalking,  trolling, and hate speech \cite{wiegand2018overview}\cite{zhang2018detecting}. During emergency and crises, these anti-social behaviours are stirring up immensely and thus deliberately or unintentionally create a hazardous effect towards a community. COVID-19 pandemic is one such situation that has changed people's lifestyles by confining them to homes and engaging in spending more time on social media. As a result, many online social media users post
hostile (such as fake, offensive, defame) contents by crossing the line defined by constitutional rights. Moreover, hostile posts on COVID-19 is a matter of concern as it can impel people to take extreme actions by believing the post is real. In order to combat the hostile contents, development of an automated system is utmost important. To address the issue, the critical contributions of this work illustrates in the following:

\begin{itemize}
    \item [$\bullet$] Develop various machine learning and deep learning-based models to detect hostile texts in social media.
    \item [$\bullet$] Present performance analysis and qualitative error analysis of the system.
\end{itemize}

The rest of the paper organized as follows: the related work discussed in Section 2. Problem definition and brief analysis of the dataset presented in Section 3. Section 4 described the methods used to develop the system. Findings and the results of the errors analysis presented in Section 5. Finally, conclusion and future direction are given in Section 6. 

\section{Related Work}
Identification and classification of hostile contents have become a prominent research issue in recent years. Various machine and deep learning approaches have achieved reasonable accuracy in solving various NLP tasks such as fake news, hate speech and abusive language detection. Saroj et al. \cite{saroj2020irlab} used SVM classifier along with tf-idf feature to identify the hate speech and offensive languages. They obtained the macro $f_1$ score of 78\% and 72\% respectively for Arabic and Greek dataset. Ibrohim et al. \cite{ibrohim2018dataset} exploited a combination of n-gram tf-idf features with SVM and Naive Bayes (NB) classifier to detect the abusive language. Their approach achieved the highest $f_1$ score of 86.43\% for the NB classifier with unigram feature. A machine learning-based approach used by Gaydhani et al. \cite{gaydhani2018detecting} to classify the hate speech and offensive language on Twitter. This work used tf-idf for varying n-gram ranges and experimented with multiple classifiers such as logistic regression (LR) and SVM. Ibrohim et al. \cite{ibrohim2019multi} used a machine learning approach with problem transformation methods for classifying the multi-label hate speech and abusive content written in the Indonesian language. Authors exploited various transformation techniques such as binary relevance (BR), label powerset (LP), and classifier chain (CC). They obtained the highest accuracy 76.16\% by using word unigram feature and an ensemble approach with LP data transformation method. To detect offensive language, an LSTM and word embedding based technique is proposed by Goel et al. \cite{goel2019usf}. Sadiq et al. \cite{sadiq2020aggression} employed combination of CNN and BiLSTM method to identify aggression on Twitter. In order to detect the fake news, a hybrid convolutional neural network (CNN) is developed by Wang et al. \cite{wang2017liar}. This work also exploited BiLSTM, LR and SVM techniques for the classification. 

\section{Task Description}
The CONSTRAINT shared task comprises of two tasks: task-A and task-B. The task definition with various class labels and brief analysis of the dataset is described in the following subsections.

\subsection{Task Definition}
The goal of task-A is to identify whether a tweet contains real or fake information. The tweets are related to the Covid-19 pandemic and written in English. In task-B, we have to perform multi-label multi-class classification on five hostile dimensions such as fake news, hate speech, offensive, defamation and non-hostile. The organizers\footnote{https://constraint-shared-task-2021.github.io/} culled hostile posts in Hindi Devanagari script from Facebook and Twitter. To better understand the task, it is monumental to have a clear idea of the class labels. The organizers \cite{patwa2020fighting} \cite{bhardwaj2020hostility} have defined various hostile and fake classes as illustrates in the following:
\begin{itemize}
    \item [$\bullet$] \textbf{Fake:} Articles, posts and tweets provide information or make claims which are verified not to be true.
    \item [$\bullet$] \textbf{Real:} The articles, posts and tweets which provided verified information and make authentic claims.
    \item [$\bullet$] \textbf{Hate speech:} Post having the malicious intention of spreading hate and violence against specific group or person based on some specific characteristics such as religious beliefs, ethnicity, and race.
    \item [$\bullet$] \textbf{Offensive:} A post contains vulgar, rude, impolite and obscene languages to insult a targeted individual or a group.
    \item [$\bullet$] \textbf{Defamation:} Posts spread misinformation against a group or individuals which aim to damage their social identity publicly.
    \item [$\bullet$] \textbf{Non-hostile:} Posts without any hostility.
\end{itemize}

\subsection{Dataset Analysis}
Classifier models have been trained and tested with the dataset conferred by the organizers. A validation set is utilized to tune the model parameters and settle the optimal hyperparameter combination. There are two classes in task-A, and we have to deal with five overlapping classes in task-B. The number of instances used to train, validate and test the models summarized in table \ref{dataset}.

To get the useful insights, we investigated the train set. Statistics of the train set exhibited in table \ref{data-summary}. From the distribution, it observed that the training set is highly imbalanced for both tasks. In task-A, the real class has total 100k words, while the fake class has only 64k words. Although there is a considerable difference in the number of total words, the number of unique words in both classes is approximately identical. That means words in real class are more frequent than words of fake class. In task-B, the total words of the non-hostile class are about four times as much as the defame class. In average, the fake class has a maximum, and non-hostile class has the minimum number of words per texts.

\renewcommand{\arraystretch}{1.1}
\begin{table}[h!]
\begin{center}
\begin{tabular}{L{1.5cm}C{1.2cm}C{1.2cm}|C{1.2cm}C{1.2cm}C{1.2cm}C{1.2cm}C{1.2cm}}

   & \multicolumn{2}{c}{Task-A} & \multicolumn{5}{c}{Task-B} \\
\hline
   & Real & Fake & Defame & Fake & Hate & Offense & Non-Hostile\\
 \hline 
 Train & 3360 & 3060 & 564 & 1144 & 792 & 742 & 3050\\
 Validation & 1120 &1020 & 77 & 160 & 103 & 110 & 435\\
 Test & 1120 & 1020 &169 &334 & 237 & 219 & 873\\
 \hline
\end{tabular}
\vspace{0.2cm}
\caption{Number of instances in train, validation and test set for each task}
\label{dataset}
\end{center}
\end{table}

\begin{table}[h!]
\begin{center}
\begin{tabular}{L{1.5cm} L{1.8cm}  C{1.5cm} C{1.5cm}C{2cm} C{2cm}}
\hline
   & Class & Total words & Unique words & Max text length (words) & Avg. no. of words per texts \\
\hline
 \multirow{2}{*}{Task-A}  & Real& 102100 & 10029 & 58 & 30.39\\
&Fake & 64929 & 9980 & 1409 & 21.22  \\
 \hline
\multirow{5}{*}{Task-B}&Defame &17287 & 4051 & 69 & 30.65\\
    &Fake & 40265 & 7129 & 403 & 35.22 \\
    &Hate & 25982 & 5101 & 116 & 32.80  \\
  & Offense & 21520 & 4624 & 123 & 29.00  \\
 &Non-Hostile & 69481 & 9485 & 60 & 22.78\\
 \hline 
\end{tabular}
\vspace{0.2cm}
\caption{Training set statistics}
\label{data-summary}
\end{center}
\end{table}

\begin{figure}[h!]
\centering
\begin{subfigure}[h!]{0.49\textwidth}
\includegraphics[height=3.7cm, width=\textwidth]{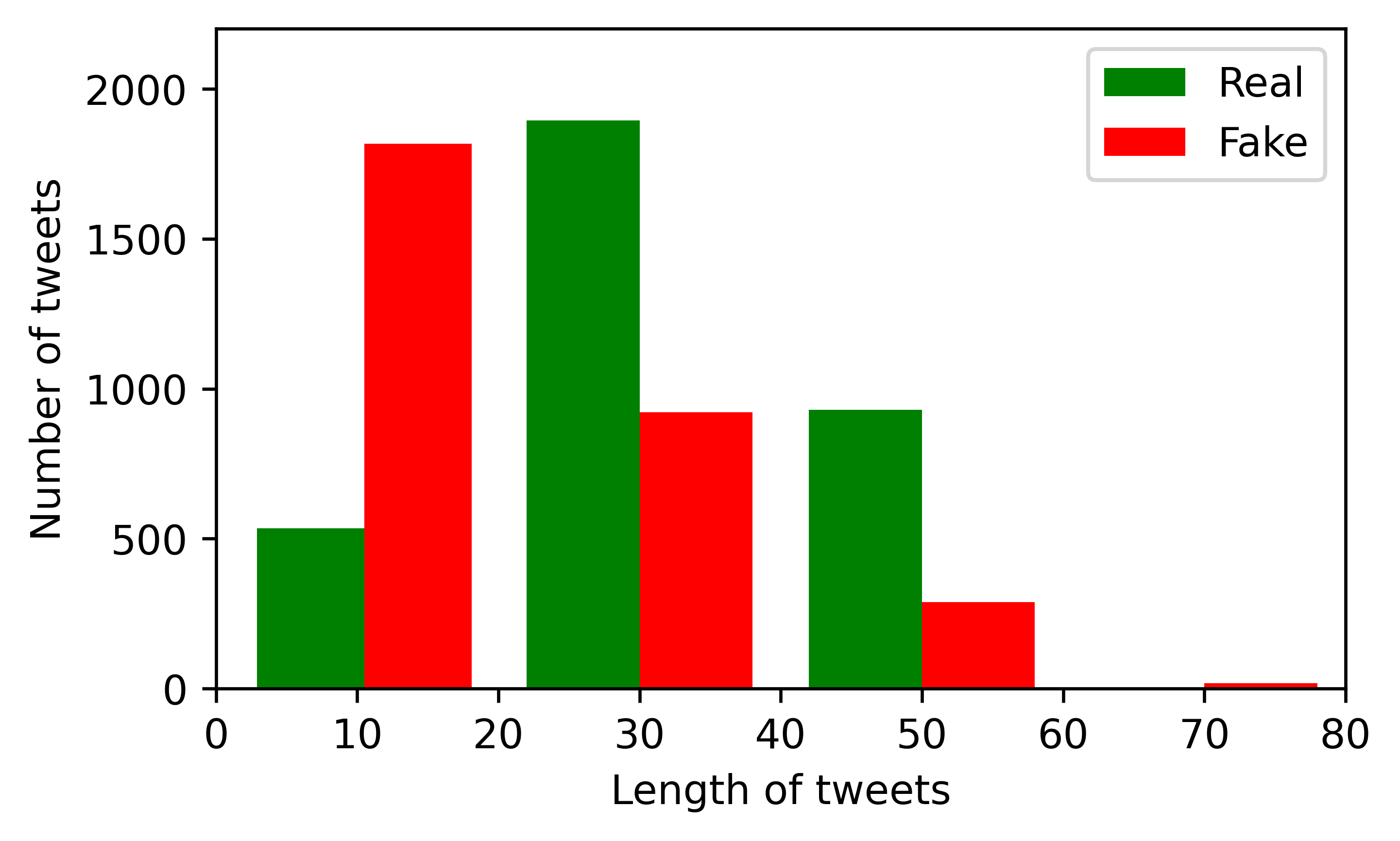}
\caption{Task-A}
\label{1a}
\end{subfigure}
\begin{subfigure}[h!]{0.49\textwidth}
\includegraphics[height=3.7cm, width=\textwidth]{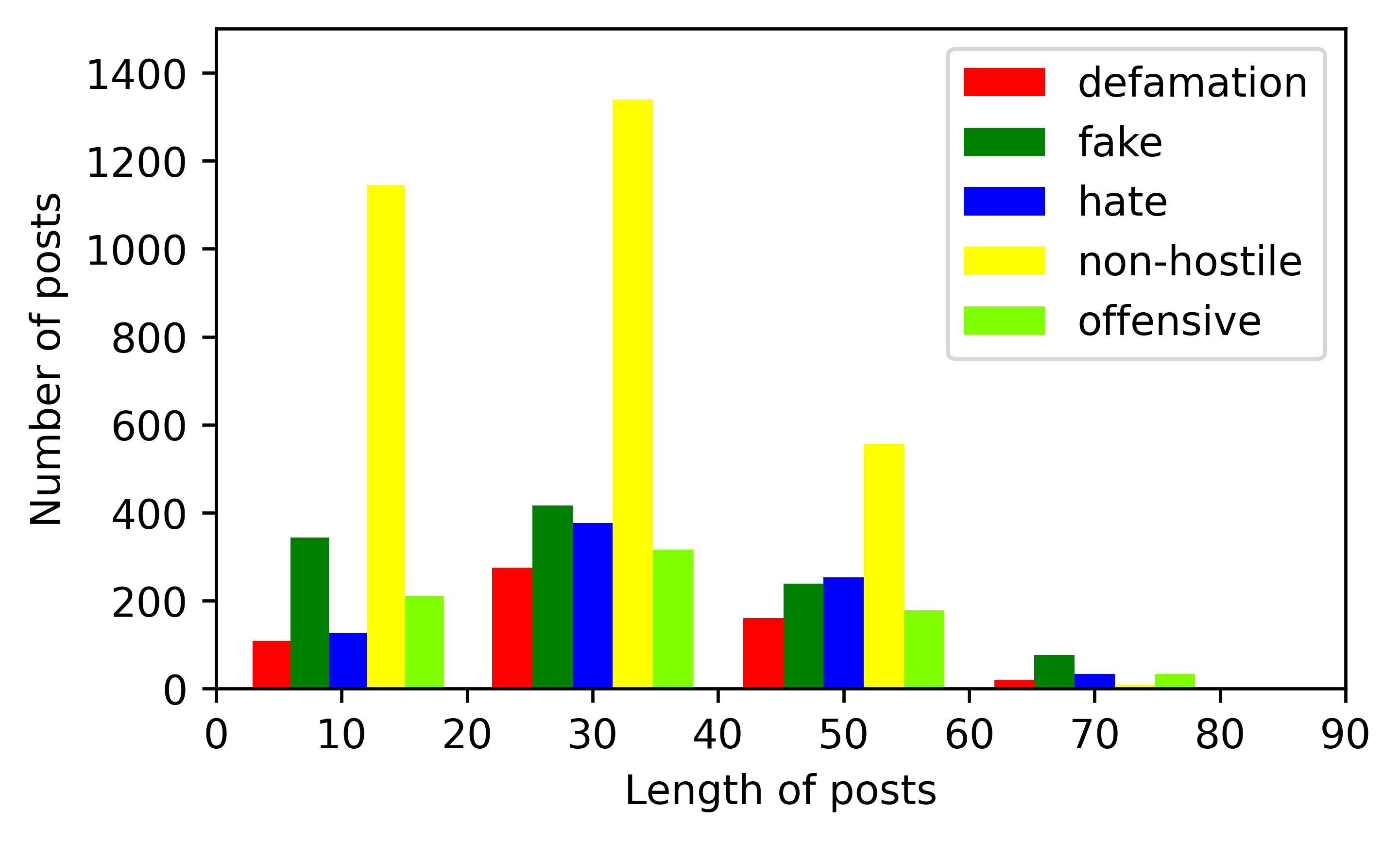}
\caption{Task-B}
\label{1b}
\end{subfigure}
\caption{Number of tweets/posts in varying length distribution scenario for different classes in training set}
\label{fig1}
\end{figure}

Figure \ref{fig1} depicts the number of texts fall in various length range. It is observed that the fake tweets length is relatively shorter than the length of real tweets. Approximately 2000 fake tweets have a length of less than 20. In contrast, more than 2800 tweets of the real class have higher than 20 words. Only a fraction of tweets has more than 60 words. Meanwhile in task-B, non-hostile class dominates in every length distribution. This difference occurs because the number of instances in non-hostile class is higher compare to other classes. Length of most of the posts is between range 20 to 40. Analysis of the training set performed after removing the punctuation, numbers and other unwanted characters.

\section{System Overview}
\begin{figure}[h!]
\centering
\includegraphics[width=\textwidth]{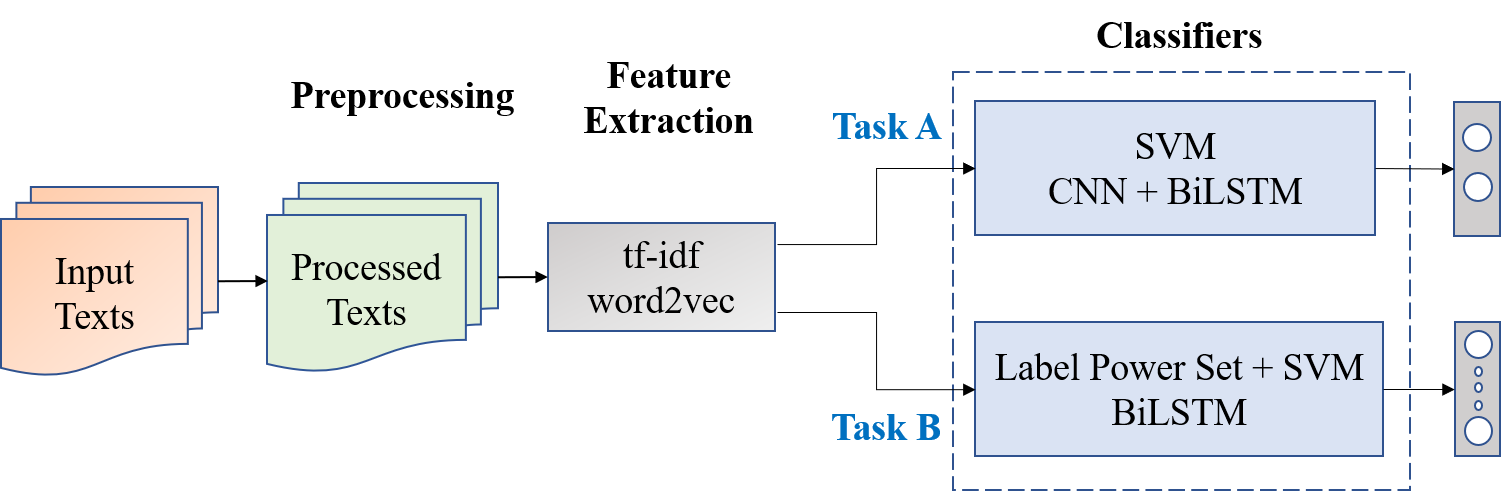}
\caption{Schematic view of the hostile text detection}
\label{architecture}
\end{figure}

Figure \ref{architecture} presents the schematic diagram of our system, which has three major phases: preprocessing, feature extraction and classification. 

\subsection{Preprocessing and Feature Extraction}
Raw input texts are processed in the preprocessing phase. All the flawed characters, numbers, emojis and punctuation's are discarded from the texts. For both tasks, identical preprocessing techniques are used. For deep learning methods, texts converted into fixed-length numeric sequences using `tokenizer' and `pad\_sequence' methods of Keras\footnote{https://keras.io/api/preprocessing/text/}. After preprocessing, features are extracted to train ML and DL models. N-gram features of the texts extracted with tf-idf technique \cite{tokunaga1994text}. Although tf-idf is a useful feature extraction technique, it could not capture the word's semantic meaning in a text. To tackle this, we used the Word2Vec word embedding technique \cite{mikolov2013distributed} that map words into a dense feature vector.  Furthermore, pre-trained word vectors \cite{grave2018learning} are also explored to investigate the models' effectiveness. These features are employed on different methods to perform the tasks. 

\subsection{Methodology}
We have employed different machine learning (ML) and deep learning (DL) approaches to crack the tasks. In this section, we describe the methods used to develop classifier models for each task. 
\subsubsection{task A:} is a binary classification problem where we have to detect whether a tweet is fake or real.
\begin{itemize}
    \item [$\bullet$] \textbf{ML approach:} SVM is implemented by both `linear' and `rbf' kernel along with tf-idf feature extraction technique. Combination of unigram and bigram feature is utilized. Value of `C' and other parameters is fixed by trial and error approach. Furthermore, SVM is applied on top of word embedding features as well. Embedding vectors are generated by varying window size, embedding dimension and other influential parameters. Other ML methods (logistic regression, decision tree, random forest) are also exploited. However, the performance of SVM is superior compare to other techniques.
    \item [$\bullet$] \textbf{DL approach:} We have implemented convolution neural network (CNN), bidirectional long short term memory (BiLSTM) and combined (CNN+ BiLSTM) network. For experimentation, network architectures are changed by varying number layers, number of neurons, dropout rate, learning rate, and other hyperparameters. In CNN, we used 64 convolution filters with kernel size $5\times 5$ and pooling window size $1\times 5$. BiLSTM network comprises of 32 bidirectional cells with dropout rate 0.2. In the combined model, these two networks are sequentially added one after another. 
\end{itemize}
\subsubsection{Task B:} is a multi-label multi-class problem where our aim is to identify different hostile dimensions .

\begin{itemize}
    \item [$\bullet$] \textbf{ML approach: } We used label powerset (LP) \cite{zhang2013review} along with support vector machine (LPSVM) classifier to develop our model. LP generates a new label for each label combination in the training corpus and thus transforming a multi-label problem into a multi-class problem \cite{tsoumakas2007random}. At first, we use tf-idf feature values of n-gram (1, 3) with linear SVM where $C = 1.7$. To reduce computation complexity, features that appear in less than five documents are discarded. In other approaches, unigram and n-gram (1,2) features are applied along with LPSVM, where the value of `C' is taken as 3.8 and 0.7, respectively. In both cases, the linear kernel is used.
    
    \item [$\bullet$] \textbf{DL approach:} BiLSTM network is employed with Word2Vec embedding technique to capture the sequential and semantic features of the texts. To get the embedding vectors, we use entire training corpus with embedding dimension $64$. These features propagated to the LSTM layer consisting of 32 bidirectional cells. The BiLSTM layer's output transferred to the dense layer having nodes equal to the number of classes ($5$) where sigmoid activation function is used. A dropout layer with the rate of $0.1$ is introduced between the LSTM and dense layer to mitigate the over-fitting. We use `binary crossentropy' as loss function and `adam' as optimizer with learning rate $0.01$.  
\end{itemize}

\section{Experiments}
Experiments were conducted on Google co-laboratory platform along with Python = 3.6.9. Machine learning models are developed by using scikit-learn = 0.22.2 packages. Besides, Keras = 2.4.0 with TensorFlow = 2.3.0 framework is chosen to implement the deep learning models. Models are developed over train data, whereas the validation set is used for tweaking the model parameters. Several measures such as accuracy (A), precision (P), recall (R) and $f_1$-score (F) are chosen to evaluate the models. Moreover, coarse-grained and fine-grained scores used to compare the performance of the models.

\subsection{Results}
In task-A, the superiority of the model is determined based on the weighted $f_1$ score. On the other hand, coarse-grained (CG) and fine-grained (FG) $f_1$-score is used to find the best task-B model. Table \ref{evataskA} presents the evaluation results of task-A on the test set. We have reported the outcomes for four models. The results revealed that the combination of CNN and BiLSTM approach achieved $f_1$ score of 92.01\%. In contrast, two models with the combination of SVM and Word2Vec obtained slightly higher $f_1$ score of 92.66\% and 92.94\% respectively. However, SVM with a tf-idf approach shows about 2\% rise and exceeds all the approaches by achieving the highest $f_1$ score (94.39\%). Our best model performance lags almost 4\% compared to the best result ($f_1$ score = 98.69\%) obtained in task-A.

\begin{table}[h!]
\centering
\begin{tabular}{C{5cm} C{1cm} C{1cm}C{1cm}C{1cm}}
\hline
Method & A & P & R & F \\
\hline
CNN + BiLSTM & 92.01 & 92.01 & 92.01 & 92.01\\
SVM + TF-idf & 94.35 & 94.42 & 94.39 & \textbf{94.39}\\
SVM + Word2Vec (ED=200) & 92.66 & 92.67 & 92.66 & 92.66\\
SVM + Word2Vec (ED=150) & 92.94 &  92.94 & 92.94 & 92.94\\
\hline
Best & 98.69 & 98.69 & 98.69 & 98.69\\
\hline
\end{tabular}
\vspace{0.2cm}
\caption{Evaluation results of task-A on test set. Here A, P, R, F denotes accuracy, precision, recall, weighted $f_1$ score respectively and ED indicates embedding dimension}
\label{evataskA}
\end{table}

\vspace{-0.8cm}
Evaluation results of task-B on the test set are presented in table \ref{evataskB}. With BiLSTM and word embedding features, we obtained CG and FG $f_1$ score of 83.37\% and 52.80\% respectively. After employing LPSVM with unigram tf-idf features, it shows a slight rise in CG $f_1$ score (84.10\%). However, FG $f_1$ score drop approximately 3\% amounting to 49.12\%. For n-gram (1, 2) with LPSVM, We observed a further rise in CG $f_1$ score (85.31\%), while FG score also increased slightly to 50.98\%. Surprisingly, we achieve the highest CG $f_1$ score of 86.03\% by varying n-gram range to (1, 3) with LPSVM. Nevertheless, FG $f_1$ score further decrease from the previous score to 50.66\%. Meanwhile, for defamation, and fake class BiLSTM + Word2Vec technique provides the highest $f_1$ score of 28.65\% and 52.06\% respectively. On the other hand, LPSVM + Ngram (1, 2) and LPSVM + Ngram (1, 3) methods give the highest $f_1$ score of 64.97\% (for fake class) and 57.91\% (for offensive class). Compared to the best CG ($f_1$ score = 97.15\%) and FG ($f_1$ score = 64.40\%) obtained in task-B, our best performing model lags approximately more than 10\% and 14\% respectively in both scores.
\begin{table}[h!]
\centering
\begin{tabular}{C{4cm} C{1cm} C{1.3cm}C{1cm}C{1cm}C{1.3cm}C{1cm}}
\hline
Method & CG & Defame & Fake & Hate & Offense & FG \\
\hline

BiLSTM + Word2Vec &  83.37 & 28.65 & 63.63 & 52.06 & 55.72 & 52.80\\
LPSVM + Unigram & 84.10 & 25.81 & 61.30 & 44.39 & 53.59 & 49.12\\
LPSVM + Ngram (1,2) & 85.31 & 27.59 & 64.97 & 47.21 & 51.72 & 50.98\\
LPSVM + Ngram (1,3) & \textbf{86.03} & 21.74 & 63.33 & 46.67 & 57.91 & 50.66\\
\hline
Best & 97.15 & 45.52 & 82.44 & 59.78 & 62.44 & 64.40  \\
\hline
\end{tabular}
\vspace{0.2cm}
\caption{Evaluation results of Task-B on the test set. All values presented in $f_1$ score and CG, FG denotes coarse-grained, and fine-grained $f_1$ scores}
\label{evataskB}
\end{table}

\vspace{-1.3cm}
\subsection{Error Analysis}
The results revealed that SVM+tf-idf and LPSVM+Ngram (1, 3) are the best performing models for task-A and task-B. To get more insights, quantitative error analysis of classification models carried out by using the confusion matrix. Tables \ref{tab5a}-\ref{tab5f} represent the confusion matrices of the classes for task-A and B.
\begin{table*}[h!]
           \centering
           \captionsetup[subtable]{position = below}
           \begin{subtable}{0.4\linewidth}
               \centering
               \begin{tabular}{C{1.5cm}| C{1.5cm} C{1.5cm}}
                \hline
                 Class & Real & Fake  \\
                \hline
                Real & $1047$ & $73$   \\
                Fake & $47$ & $973$   \\
                \hline 
                \end{tabular}
               \caption{Task-A}
               \label{tab5a}
           \end{subtable}%
           \hspace*{2em}
           \begin{subtable}{0.4\linewidth}
               \centering
              \begin{tabular}{C{1.5cm}|   C{1.5cm}C{1.5cm}}
                \hline
                Class & Defame & Other  \\
                \hline
                Defame & $25$ & $144$   \\
                Other & $42$ & $1442$   \\
                \hline 
                \end{tabular}
                 \caption{Task-B:Defame}
                \label{performance}
           \end{subtable}
           \begin{subtable}{0.4\linewidth}
               \centering
              \begin{tabular}{C{1.5cm}|   C{1.5cm}C{1.5cm}}
                \hline
                Class & Fake & Other  \\
                \hline
                Fake & $190$ & $144$   \\
                Other & $116$ & $1203$   \\
                \hline 
                \end{tabular}
                
                \caption{Task-B:Fake}
                \label{performance}
           \end{subtable}
           \hspace*{2em}
           \begin{subtable}{0.4\linewidth}
               \centering
              \begin{tabular}{C{1.5cm} |   C{1.5cm}C{1.5cm}}
                \hline
                Class & Hate & Other  \\
                \hline
                Hate & $91$ & $143$   \\
                Other & $75$ & $13442$   \\
                \hline 
                \end{tabular}
                
                \caption{Task-B:Hate}
                \label{performance}
           \end{subtable}
           
           \begin{subtable}{0.4\linewidth}
               \centering
              \begin{tabular}{C{1.5cm} |   C{1.5cm}C{1.5cm}}
                \hline
                Class & NH & Other  \\
                \hline
                NH & $814$ & $59$   \\
                Other & $170$ & $610$   \\
                \hline 
                \end{tabular}
         \caption{Task-B:Non-Hostile}
                \label{performance}
           \end{subtable}
           \hspace*{2em}
           \begin{subtable}{0.4\linewidth}
               \centering
              \begin{tabular}{C{1.5cm}|   C{1.5cm}C{1.5cm}}
                \hline
                Class & Offense & Other  \\
                \hline
                Offense & $108$ & $111$   \\
                Other & $49$ & $1385$   \\
                \hline 
                \end{tabular}
     \caption{Task-B:Offensive}
               \label{tab5f}
           \end{subtable}
            \caption{Confusion matrix for SVM + tf-idf (task-A) and LPSVM + Ngram(1,3) (task-B)}
\end{table*}
\vspace{-1.5cm}
\subsubsection{Quantitative analysis:} 
Table \ref{tab5a} indicates that a total of 73 real tweets wrongly identified as fake and the system marked 47 fake tweets as real. The false-negative rate in defame, hate, and offensive classes are very high. Only 25, 91 and 108 posts are correctly classified among 169, 134 and 219 posts respectively for these classes. Majority of these tweets are wrongly classified as either fake or non-hostile. In contrast, the false positive rate is high for fake and non-hostile classes. A total of 116 posts are incorrectly classified as fake while the model could not identify 144 actual fake posts. The model also even suffered to differentiate the hostile and non-hostile posts and thus misclassified 170 hostile posts. Dataset imbalance might be the reason behind this poor performance. The model trained with many fakes, and non-hostile class instances compare to other hostile classes. Increasing the number of training examples will undoubtedly help the model to perform better.
\begin{figure}[h!]
\centering
\includegraphics[width=\textwidth]{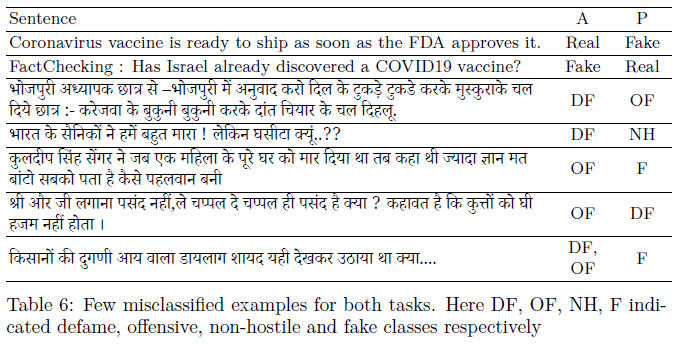}
\label{table11}
\end{figure}
\vspace{-1.5cm}
\subsubsection{Qualitative analysis:} Some misclassified examples with there actual (A) and predicted label (P) presented in table 6. After giving a closer look at the examples, we discover some interesting facts. The error occurs due to some standard frequent terms such as `covid19', `coronavirus', `modi', and `vaccine' that exist in real and fake tweets. These influential words make it difficult to differentiate between fake and real. We also notice that fake news claims unverified facts using responsible agencies, i.e. FDA, and WHO. It is challenging to verify such claims and make a proper prediction. In the case of hostility detection, some posts inherently express hostility which is very arduous to identify from the surface level analysis without understanding the context. Moreover, due to the overlapping characteristics of hostile dimensions, confusion mostly occurred when separating the defame, hate and offensive posts. Analyzing the context of the posts might help to develop more successful models. 

\subsection{Discussion}
Surprisingly, machine learning models have performed better than deep learning models in the CONSTRAINT shared tasks. As deep learning techniques are the data-driven method, lack of training examples in a few classes might be a reason for this peculiar behaviour. In order to handle this issue, pretrained word embedding can be utilized. However, noticeable change in the performance was not observed for example, combined (CNN+BiLSTM) model obtained 92.43\% weighted $f_1$ score with pretrained word vectors which is lower than the SVM (94.39\%). After analysis, we realized that large pretrained language models might help the system make predictions more accurately. Thus, after getting the actual text labels at the end of the shared task, we applied the BERT model \cite{devlin2018bert} and notice an astonishing rise in accuracy for both tasks. In task-A weighted $f_1$ score increased from  0.94 to 0.98 and coarse-grained $f_1$ score incremented from 0.86 to 0.97. Table \ref{bert} shows the outcomes of the BERT model.
\begin{table}[h!]
\begin{center}
\begin{tabular}{L{1.5cm}|C{1.2cm}|C{1.2cm}C{1.2cm}C{1.2cm}C{1.2cm}C{1.2cm}C{1.2cm}}

   & {Task-A} & \multicolumn{5}{c}{Task-B} \\
\hline
   & $f_1$ score & Coarse grained & Defame & Fake & Hate & Offense \\
 \hline 
BERT & 0.98 & 0.97 & 0.38 & 0.79 & 0.46 & 0.54\\
 \hline
\end{tabular}
\vspace{0.2cm}
\caption{Results obtained by the BERT model on task-A and task-B}
\label{bert}
\end{center}
\end{table}
\vspace{-1.2cm}

\section{Conclusion}
This paper presents the system description with detailed results and error analysis developed in the CONSTRAINT 2021 shared task. Various learning techniques have explored with tf-idf feature extraction, and Word2Vec embedding technique to accomplish the tasks A and B. Results shows that SVM with tf-idf achieved the highest of 94.39\% $f_1$ scores for the task-A and LPSVM with n-gram (1, 3) obtained the highest of 86.03\% $f_1$ scores for the task-B. However, the BERT pre-trained model provided the 98\% accuracy in task-A and 97\% accuracy in task-B. Since CNN and BiLSTM did not achieve satisfactory accuracy, it will be interesting to see how they perform after applying ensemble technique or adding attention layer. Increasing the number of posts in hostile classes can help to improve the performance of the models. These issues will address in future work.

%
%
\bibliographystyle{splncs03_unsrt.bst}
\bibliography{samplepaper}


\end{document}